\documentclass{article}

\PassOptionsToPackage{numbers, compress}{natbib}

\usepackage[preprint]{neurips_2020}

\usepackage[utf8]{inputenc} 
\usepackage[T1]{fontenc}    
\usepackage{hyperref}       
\usepackage{url}            
\usepackage{booktabs}       
\usepackage{amsfonts}       
\usepackage{nicefrac}       
\usepackage{microtype}      

\usepackage[utf8]{inputenc}
\usepackage[font=normal,skip=0pt]{caption}
\usepackage{graphicx}
\usepackage{amsmath}
\usepackage{amsthm}
\usepackage{algorithm}
\usepackage{algorithmic}
\usepackage{stmaryrd}
\usepackage{mathtools}
\usepackage{xcolor}
\usepackage[font=scriptsize, skip=1pt]{subcaption}
\usepackage{xspace}
\usepackage{dsfont}
\usepackage{enumitem}
\usepackage{amssymb}
\usepackage{multicol}
\usepackage{wrapfig}
\usepackage{bm}
\usepackage{pifont}
\usepackage{tikz}

\usetikzlibrary{shapes}
\usetikzlibrary{shapes.geometric}

\newcommand{\circleB}{\protect\tikz\node[circle ,fill={rgb,255:red,61;green,133;blue,198},draw,scale=0.6]{};\xspace}
\newcommand{\starEnd}{\tikz\node[star   ,fill={rgb,255:red,255;green,255;blue,0},draw,scale=0.4,star point ratio=2.25]{~};\xspace}

\newcommand{\algskip}{\par\hskip\algorithmicindent}

\newcommand{\iil}{\textsc{I2L}\xspace}
\newcommand{\ail}{\textsc{AIL}\xspace}

\newcommand{\method}{\textsc{APIL}\xspace}
\newcommand{\intrinsic}{\textsc{IntrUn}\space}
\newcommand{\extrinsic}{\textsc{ExtrUn}\space}
\newcommand{\behave}{\textsc{BehvUn}\space}
\newcommand{\bc}{\textsc{BC}\xspace}
\newcommand{\dgger}{\textsc{DAgger}\xspace}
\newcommand{\errpred}{\textsc{ErrPred}\space}

\newcommand{\detm}{\textsc{Detm}\xspace}
\newcommand{\rand}{\textsc{Rand}\xspace}
\newcommand{\tworand}{\textsc{TwoRand}\xspace}
\newcommand{\twodifdetm}{\textsc{TwoDifDetm}\xspace}

\newcommand{\toynav}{\textsc{Grid}\xspace}
\newcommand{\rr}{\textsc{R2R}\xspace}

\newcommand{\exepol}{\hat{\pi}_{\textrm{exe}}}
\newcommand{\exeexp}{\pi^{\star}_{\textrm{exe}}}
\newcommand{\askpol}{\hat{\pi}_{\textrm{ask}}}
\newcommand{\askexp}{\pi^{\star}_{\textrm{ask}}}
\newcommand{\exea}{\hat{a}^{\textrm{exe}}_t}
\newcommand{\aska}{\hat{a}^{\textrm{ask}}_t}
\newcommand{\exearef}{a^{\textrm{exe}\star}_t}
\newcommand{\askaref}{a^{\textrm{ask}\star}_t}
\newcommand{\jointpol}{\hat{\pi}}

\newcommand{\traj}{\hat{\tau}}
\newcommand{\proset}{S^{\odot}_{\jointpol}}
\newcommand{\appoxproset}{S^{\odot}_{\traj}}

\newcommand{\texttts}[1]{{\texttt{#1}}}

\def\equationautorefname~#1\null{Eq~#1\null}
\def\algorithmautorefname~#1\null{Alg~#1\null}
\def\tableautorefname~#1\null{Tab~#1\null}
\def\figureautorefname~#1\null{Fig~#1\null}
\def\sectionautorefname~#1\null{Sec~#1\null}

\title{Active Imitation Learning from Multiple Non-Deterministic Teachers: \\Formulation, Challenges, and Algorithms}

\newcommand{\idcs}{${}^\diamondsuit$}
\newcommand{\idmsr}{${}^\heartsuit$}

\author{Khanh Nguyen\idcs \and Hal Daum{\'e} III\idcs\idmsr \\
  University of Maryland, College Park\idcs, Microsoft Research, New York\idmsr\\
  \tt{ \{kxnguyen,hal\}@umiacs.umd.edu }
 }

\author{%
  Khanh Nguyen\idcs \qquad  Hal Daum{\'e} III\idcs\idmsr \\
  University of Maryland, College Park\idcs, Microsoft Research, New York\idmsr\\
  \texttt{ \{kxnguyen,hal3\}@umd.edu }
}

\begin{document}

\maketitle

\begin{abstract}
  We formulate the problem of learning to imitate multiple, non-deterministic teachers with minimal interaction cost.
Rather than learning a specific policy as in standard imitation learning, the goal in this problem is to learn a distribution over a policy space.
We first present a general framework that efficiently models and estimates such a distribution by learning continuous representations of the teacher policies.
Next, we develop \textit{Active Performance-Based Imitation Learning} (\method), an active learning algorithm for reducing the learner-teacher interaction cost in this framework.
By making query decisions based on predictions of future progress, our algorithm avoids the pitfalls of traditional uncertainty-based approaches in the face of teacher behavioral uncertainty. 
Results on both toy and photo-realistic navigation tasks show that \method significantly reduces the numbers of interactions with teachers without compromising on performance.  
Moreover, it is robust to various degrees of teacher behavioral uncertainty.
\end{abstract}

\section{Introduction}
Teaching agents to communicate with humans to accomplish complex tasks is a long-standing goal in machine learning. 
Interactive Imitation Learning (\iil) \cite{daume2009search,ross2010efficient,ross2011reduction,ross2014reinforcement,chang2015learning,sharaf2017structured} studies a special case where a learning agent learns to fulfill tasks by querying and imitating decisions suggested by experts (referred to as \textit{teachers}). 
This framework has shown promising results on various research problems, demonstrating advantages against other frameworks like non-interactive imitation learning \cite{ross2011reduction} and reinforcement learning \cite{sun2017deeply}.

Unfortunately, standard \iil is notoriously query-inefficient because it queries the teacher in \textit{every} state the agent visits. 
Active Imitation Learning (\ail) \cite{judah2014active} offers a more practical framework by allowing the agent to actively decide in which states it wants to query the teacher. 
Despite extensive research efforts, \ail has mainly focused on interaction with a single, deterministic teacher. 
Nevertheless, learning multiple and non-deterministic task-solving policies is highly desirable in various scenarios.
Mastering multiple ways to accomplish a task may provide deeper insights and enhance generalizability to novel tasks.
Learning to act non-deterministically may be helpful in stochastic environments. 
In some cases, it may be not possible or too costly to designate a teacher that monitors an agent all the time. 
Being able to acquire knowledge from diverse sources enables agents to leverage additional help from undesignated teachers scattered in the environment \cite{nguyen2019hanna,nguyen2019vision,thomason2019vision}. 

While enhancing capabilities of learning agents, employing multiple, non-deterministic teachers presents new challenges in designing robust \ail algorithms.
Concretely, these algorithms must be able to learn in the face of \textit{behavioral uncertainty} that stems from each individual teacher, or from discrepancies among teachers. 
In this work, we show that behavioral uncertainty from teachers causes the agent to distribute high probabilities to multiple policies or actions, appearing highly uncertain. 
\ail algorithms that rely on uncertainty \cite{cronrath2018bagger,hanczor2018improving} or disagreement \cite{laskey2016shiv, kim2013maximum} may misunderstood this behavior as a sign of incompetence and may issue redundant queries. 
Consider a simple example where two teachers (A and B) disagree in teaching an agent how to handle a situation: A suggests the agent select action 1, whereas B favors action 2. 
Suppose both actions are optimal; the teachers simply have different preferences.
If the agent manages to imitate both teachers, it will learn a \textit{mean} action distribution that is highly uncertain and incurs a large imitation error with respect to each teacher. 
Any strategy that issues queries upon high uncertainty or error will likely query the teachers indefinitely and unnecessarily.
However, if the agent learned that both action choices lead to desirable outcomes, it would proceed to take either action without querying. 
This example motivates the need for reasoning about task progress when learning in the face of teacher behavioral uncertainty. 

Our paper makes three main contributions:
\begin{enumerate}
    \item We present a general framework for imitation learning from multiple, non-deterministic teachers. Our framework models a distribution over an entire policy space by learning continuous representations of policies;
    \item Using this framework, we theoretically and empirically analyze how traditional uncertainty-based approaches to \ail are deficient when the teachers exhibit behavioral uncertainty;
    \item We develop \textit{Active Performance-Based Imitation Learning} (\method), a novel \ail algorithm that bases query decisions on predictions of future task progress. Experimental results on two navigation tasks demonstrate the efficiency and robustness of our algorithm against various types of teacher uncertainty.
\end{enumerate}

\section{Preliminaries}
\noindent\textbf{Environment and Problem}. We model an agent living in an environment with state space $S$. 
Each fully-observed state $s \in S$ captures relevant information in the environment, including the state of the agent. 
At every time step, the agent can take an action $a$ in an action space $A$ to transition to a next state.
Initially, the agent receives a task in state $s_{\textrm{start}}$. 
To execute the task, it maintains an execution policy $\hat{\pi}$, which, at time step $t$, takes as input a state $s_t$ and outputs an action distribution $\hat{\pi}(\cdot \mid s_t)$ over $A$.
We denote by $\hat{a}_t \sim \hat{\pi}(\cdot \mid s_t)$ the action selected by the agent at time step $t$.
We impose a time constraint that the agent has to terminate its execution after $T$ steps. 
We define a goal set $S_{\textrm{goal}}$ representing the set of states in which the agent is considered "successfully fulfilling" the task.
A non-negative metric $d(s)$ quantifies the ``distance" between a state $s$ and the goal set $S_{\textrm{goal}}$;
$d(s) = 0$ if and only if $s \in S_{\textrm{goal}}$.
The detailed implementation of this metric depends on the application.
For example, in robot navigation, $S_{\textrm{goal}}$ is the set of goal locations and $d(s)$ computes the shortest-path distance between the agent's current location and the goal location.

\noindent\textbf{Interactive Imitation Learning (\iil)}. In the standard \iil setup \cite{daume2009search,ross2011reduction}, the agent learns by interacting with a teacher that implements a policy $\pi^{\star}$.
Upon a request from the agent in a state $s$, the teacher references an action $a^{\star}_s \sim \pi^{\star}(\cdot \mid s)$ that the agent should take.
The agent's goal is to minimize the action-imitation loss with respect to the reference action, averaged over the states it visits
\begin{align}
    \hat{\pi} = \arg\min_{\pi} \mathbb{E}_{s \sim P
    _{\pi}}\left[ \ell(s, \pi, a^{\star}_s)  \right]
    \label{eqn:orin_il}
\end{align} where $P_{\hat{\pi}}$ is the state distribution that the agent induces by traversing in the state space using its own policy, and $\ell$ is a loss function that quantifies the difference between the decision of the agent and that of the teacher.


\noindent \textbf{Active Imitation Learning (\ail)}. In this setup, the agent maintains two policies: an execution policy $\exepol$ for performing the main task, and a query policy $\askpol$ for deciding when to query an execution teacher $\exeexp$.
The query policy maps a state to an action distribution over the query action space $A^{\textrm{ask}} = \{\texttts{continue}, \texttts{query}\}$. 
At time step $t$, the agent selects a query action $\aska \sim \askpol(\cdot \mid s)$.
If $\aska = \texttts{continue}$, the agent draws an action $\exea \sim \exepol(\cdot \mid z_t)$ from its own execution policy and executes this action without querying the teacher. 
If $\aska = \texttts{query}$, the agent queries the teacher and receives a reference action $\exearef \in A$.
In addition, we assume that, upon each query and at the end of a task episode, the teacher returns $d_t \triangleq d(s_t)$, the distance to the goal set from the current state.
These distances are crucial in our approach as we rely on them to teach the agent to be aware of its progress on the main task.
In general, every time the agent queries the teacher, it can choose to act using either the newly received reference action or its own action. 
However, in this work, we assume the agent \emph{always} uses the reference action to act. 
This choice enables safer exploration during training, resulting in higher training performance and faster convergence. 
Given a fixed query policy, the learning objective of the execution policy is
\begin{align}
    \min_{\exepol} \mathbb{E}_{s \sim P_{\hat{\pi}}, \hat{a}^{\textrm{ask}}_s \sim \askpol(\cdot \mid s)}\left[ \ell_{\textrm{NLL}}(s, \exepol, a^{\textrm{exe}\star}_s) \cdot \mathds{1}\{ \hat{a}^{\textrm{ask}}_s = \texttts{query} \} \right]
    \label{eqn:active_exepol}
\end{align} where $\hat{\pi} = \left( \exepol, \askpol \right)$, and $\mathds{1}\{.\}$ is an indicator function. Notice two differences compared to standard \iil: (a) the agent-induced visitation distribution $P_{\hat{\pi}}$ depends on both the execution and the query policies and (b) an imitation loss is only incurred when the agent queries.


\section{Active Imitation Learning with a Theory of Teaching Mind}
\label{sec:mind}
\noindent \textbf{Problem Formulation}. We model scenarios where the agent interacts with multiple, (possibly) non-deterministic teachers.
For simplicity, we assume there are $K$ teachers in the environment. 
The $k^{\textrm{th}}$ teacher implements policy $\pi^{\star}_k$.
The teacher policies are not necessarily distinct.
Upon a query from the agent, the environment selects a teacher to interact with the agent. 
The selected teacher responds to the query with a reference action $a^{\star}$ and an index $k \in \{1, \ldots, K \}$ representing its identity.
To keep our framework general, we do not impose any assumptions on the teacher-selection process or the teacher policies; they can be deterministic or non-deterministic depending on the scenario.

In this setup, it is not sufficient to learn a single teacher policy. 
To fully leverage the knowledge of the teachers, our goal is to approximate the true (state-dependent) policy distribution $p^{\star}(\pi \mid s)$ by a learned distribution $\hat{p}_{\omega}(\pi \mid s)$ (parameterized by $\omega$).
We learn a distribution over the policy space rather than the identity space to effectively characterize the disagreement among the teachers, as the latter space may contain duplicate policies.
This formulation will be helpful in the later discussion on uncertainty induced by multiple teachers.

\noindent \textbf{Modeling}. 
To construct a distribution over distinct policies, we introduce the concept of \textit{policy persona}, which uniquely characterizes a policy.
Two teachers have identical personae iff they implement the same policy. 
We model $\hat{p}_{\omega}$ as a stochastic process constituted by three components: (a) an \textit{identity distribution} $\rho_{\psi}(k \mid s)$ estimates a state-dependent distribution over teacher identities; (b) an \textit{persona model} $h_{\phi}(k)$ computes a (vector) representation of each teacher policy; and (c) a \textit{persona-conditioned policy} $\hat{\pi}_{\theta, h}(a \mid s)$ defines an execution policy parameterized by a persona representation $h$ and internal parameters $\theta$.
The process for sampling a policy from $\hat{p}_{\omega}(\cdot \mid s)$ proceeds as follows: (a) sample an identity $k \sim \rho_{\psi}(\cdot \mid s)$; (b) compute its persona representation $h = h_{\phi}(k)$; and (c) return $\hat{\pi}_{\theta, h}$ as the sample. 

\begin{figure}
\begin{minipage}[t]{0.48\textwidth}
\begin{algorithm}[H]
\algsetup{linenosize=\scriptsize}
\caption{Teacher Persona-Aware Active Imitation Learning with a fixed query policy}
\scriptsize
\label{alg:active_exe}
\begin{algorithmic}[1] 
\STATE set $s_0 \leftarrow s_{\textrm{start}}$
\FOR{$t = 0 \cdots T - 1$}
\STATE choose a query action $\aska \sim \askpol(\cdot \mid s_t)$
\IF{$\aska = \texttts{query}$}
\STATE query and receive: (a) reference action $\exearef$, (b) distance \algskip to goals $d_t$, and (c) teacher identity $k_t^{\star}$
\STATE compute persona representation $h^{\star}_t \leftarrow h_{\phi}\left(k^{\star}_t \right)$
\STATE compute policy model's loss $\ell_{\textrm{NLL}}\left(s_t, \hat{\pi}_{\theta, h^{\star}_t}, \exearef\right)$
\STATE compute identity model's loss $\ell_{\textrm{NLL}}\left(s_t, \rho_{\psi}, k_t^{\star} \right)$
\STATE use reference action to act $\exea \leftarrow \exearef$
\ELSE
\STATE approximate mean agent policy: $\exepol = \frac{1}{N} \sum_{i = 1}^N \pi_{\textrm{exe}}^{(i)}$ \algskip with $N$ samples $\pi_{\textrm{exe}}^{(i)} \sim \hat{p}(\cdot \mid s)$
\STATE sample an action to act $\exea \sim \exepol(\cdot \mid s_t)$
\ENDIF
\STATE take action $\exea$ and arrive in new state $s_{t + 1}$
\ENDFOR
\STATE receive final distance to goal set $d_T$
\STATE update $\theta, \phi$ to minimize policy model's total loss
\STATE update $\psi$ to minimize identity model's total loss
\end{algorithmic}
\end{algorithm}
\end{minipage}
\hfill
\begin{minipage}[t]{0.48\textwidth}
\begin{algorithm}[H]
\algsetup{linenosize=\scriptsize}
\scriptsize
\caption{Active Performance-Based Imitation Learning (\method)}
\label{alg:apil}
\begin{algorithmic}[1] 
\STATE execute Alg \ref{alg:active_exe} to obtain a trajectory $\traj = \left\{ \left( s_t, \exea, \aska \right)\right\}_{t = 1}^T$ and distances to goal set $\{ d_q \}_{q \in Q}$ where $Q \triangleq \{T\} \cup \{ t \in \{1, \ldots, T\} \mid \aska = \texttts{query} \}$
\STATE set $d_{\textrm{min}} \leftarrow d_T$, $progress \leftarrow \mathds{1}\{d_T \leq \epsilon \}$
\FOR{$t = (T - 1) \cdots 0$}
\IF {$\aska = \texttts{query}$}
\STATE $progress \leftarrow progress \vee \mathds{1}\{ d_t \leq \sigma \cdot d_{\textrm{min}} \}$
\STATE update $d_{\textrm{min}} \leftarrow \min\left( d_{\textrm{min}}, d_t \right)$
\ENDIF
\IF {$progress$}
\STATE $\askaref \leftarrow \texttts{continue}$
\ELSE
\STATE $\askaref \leftarrow \texttts{query}$
\ENDIF
\STATE compute imitation loss $\ell_{\textrm{NLL}}\left(s_t, \askpol, \askaref \right)$
\ENDFOR
\STATE update $\askpol$ to minimize total imitation loss
\end{algorithmic}
\vspace{0.95cm}
\end{algorithm}
\vfill
\end{minipage}
\vspace{-.6cm}
\end{figure}

\noindent\textbf{Learning}. During training, through the responses of the teachers, the agent observes the reference actions and the true identities of the teachers that recommend those actions. 
We learn the components of $\hat{p}_{\omega}$ by minimizing imitation losses with respect to the teacher responses:
\begin{align}
    \min_{\phi, \theta} \mathbb{E}_{s \sim P_{\hat{\pi}}}
    \left[ \ell_{\textrm{NLL}}\left(s, \hat{\pi}_{\theta, h_s^{\star}}, a^{\star}_s  \right)  \right] + 
    \min_{\psi} \mathbb{E}_{s \sim P_{\hat{\pi}}}
    \left[ \ell_{\textrm{NLL}}\left(s, \rho_{\psi}, k^{\star}_s \right) \right]
\end{align} where $\left(a^{\star}_s, k^{\star}_s\right)$ is the response in state $s$, $h^{\star}_s \triangleq h_{\phi}\left(k^{\star}_s\right)$ is the persona representation of teacher $k^{\star}_s$, and $\hat{\pi}(\cdot \mid s) = \mathbb{E}_{\pi \sim \hat{p}_{\omega}}\left[ \pi(\cdot \mid s) \right]$ is the mean agent policy. 

Similar to standard \iil, we let the agent act in the environment using its own (mean) policy $\hat{\pi}$, which is averaged over the predicted persona distribution.
However, when computing the action-imitation loss (the first term), we use the policy that is conditioned on the observed persona $h_s^{\star}$.
The goal here is to learn representations that capture the associations between the personae and their behaviors (e.g. persona A is frequently associated with action 1, while persona B is usually coupled with action 2).
Because the agent policies share the internal parameters $\theta$, those that exhibit similar behaviors must learn similar persona representations.
On the other hand, the second term of the objective teaches $\rho_{\psi}$ to estimate the identity distribution.

In our implementation, $\rho_{\psi}$ and $\pi_{\theta, h}$ are modeled by neural networks that output probability distributions, and $h_{\phi}$ is an embedding lookup table. 
In particular, $\pi_{\theta, h}$ is a neural network that concatenates a persona vector $h$ with a state vector $s$, and takes the result as input. 
Our framework can be easily extended to AIL (\autoref{alg:active_exe}).
Similar to the standard AIL setup, imitation losses are only incurred when the agent decides to query. 
The acting rule remains unchanged: if the agent queries, it follows the reference action $a^{\star}_s$; otherwise, it follows its own execution policy $\exepol(\cdot \mid s) \triangleq \mathbb{E}_{\pi \sim \hat{p}_{\omega}}\left[ \pi(\cdot \mid s) \right]$.

\section{Active Imitation Learning in the Face of Teacher Behavioral Uncertainty}
\label{sec:uncertainty}
\noindent\textbf{Sources of Uncertainty}. Our framework enables analyses of uncertainty of the reference actions, which we referred to as \textit{teacher behavioral uncertainty}. 
Concretely, it accounts for two sources of teacher behavioral uncertainty: \textit{intrinsic (behavioral) uncertainty}, which arises from the non-deterministic behavior of each teacher, and \textit{extrinsic (behavioral) uncertainty}, which stems from the discrepancies among the teachers.
To mathematically define uncertainty, we use variance for measuring uncertainty of continuous distributions, and use Shannon entropy \cite{shannon1948mathematical} for discrete distributions. 
We apply the law of total variance and the definition of mutual information to characterize how the two sources of behavioral uncertainty contributes to the total behavioral uncertainty:
\begin{align}
    \textrm{Continuous: }\underbrace{\textrm{Var}\left[a^{\star} \mid  s\right]}_{\textrm{total behavioral uncertainty}} &= \underbrace{\mathbb{E}_{\pi^{\star} \sim p^{\star}}\left[ \textrm{Var}\left[a^{\star} \mid \pi^{\star}, s \right]\right]}_{\textrm{intrinsic behavioral uncertainty}} + \underbrace{\textrm{Var}_{\pi^{\star} \sim p^{\star}} \left[ \mathbb{E}\left[ a^{\star} \mid \pi^{\star}, s \right] \right]}_{\textrm{extrinsic behavioral uncertainty}} \\
    \textrm{Discrete: }\underbrace{\mathbb{H}\left[ a^{\star} \mid s \right]}_{\textrm{total behavioral uncertainty}} &=
    \underbrace{\mathbb{E}_{\pi^{\star} \sim p^{\star}}\left[ \mathbb{H}\left[ a^{\star} \mid \pi^{\star}, s \right] \right]}_{\textrm{intrinsic behavioral uncertainty}} + 
    \underbrace{\mathbb{I}_{\pi^{\star} \sim p^{\star}}\left[ a^{\star}, \pi^{\star} \mid s \right]}_{\textrm{extrinsic behavioral uncertainty}} 
\end{align} where $\mathbb{H}$ and $\mathbb{I}$ are the Shannon entropy and mutual information, respectively. 
The mathematical definitions agree with our intuitions, showing that intrinsic uncertainty is related to the scatteredness of the teachers' output distributions $\pi^{\star}(a \mid s)$, whereas extrinsic uncertainty correlates with the scatteredness of the teacher distribution $p^{\star}(\pi \mid s)$.\footnote{By definition, the mutual information between $a^{\star}$ and $\pi^{\star}$ measures how much knowing about $a^{\star}$ reduces uncertainty about $\pi^{\star}$. High mutual information indicates that observing an action reveals a lot about the identity of a teacher. Thus, the teachers must be very distinct from one another.}
Alternatively, we can measure the \textit{agent behavioral uncertainty} (i.e., uncertainty of the agent's actions) by replacing $p^{\star}$ with $\hat{p}_{\omega}$ in the definitions.
It can serve as an approximation of the unknown teacher behavioral uncertainty.

\noindent\textbf{Pitfalls of Uncertainty-Based AIL}.
Uncertainty-based methods (e.g. \cite{gal2016dropout}) are dominant approaches to active learning. 
At a high level, these methods employ a simple heuristic that queries the teachers when certain estimations of the agent's uncertainty exceed a threshold: $\askpol(\texttts{query} \mid s) = \mathds{1}\{ u(s) > \tau \}$, where $u(.)$ estimates some type of uncertainty, and $\tau$ is a (usually small) constant. 

When there is only a single, deterministic teacher, implementing this heuristic with any type of behavioral uncertainty defined above (intrinsic, extrinsic, or total) is a well-motivated approach. 
As the teacher behavioral uncertainty is zero in this case, pushing the agent behavioral uncertainty below a small threshold helps aligning behavior of the agent with that of the teacher. 
In contrast, when there are multiple, non-deterministic teachers, the teacher behavioral uncertainty becomes a non-zero, unknown quantity.
Because this quantity is non-zero, there is no motivation for forcing the agent behavioral uncertainty to be small; and because the quantity is unknown, it is unclear what threshold $\tau$ the agent should drive its behavioral uncertainty below in order to match the teachers' behavior.

Using model uncertainty (a.k.a. epistemic uncertainty) seems to be a more reliable approach.  
Model uncertainty stems from lack of knowledge and generally decreases through learning, regardless of the magnitude of the teacher behavioral uncertainty.\footnote{Teacher behavioral uncertainty can be categorized as a type of \textit{aleatory uncertainty}, which comes from the data-generating process and is orthogonal to model (or epistemic) uncertainty.}
Capturing model uncertainty requires taking a Bayesian approach \cite{gal2017deep,houlsby2011bayesian,depeweg2017uncertainty}, modeling a posterior distribution $\hat{q}(\omega \mid D)$ over all model parameters conditioned on a dataset $D$ containing all observed interactions with the teachers. 
Model uncertainty can be computed in our framework by subtracting the agent's behavioral uncertainty from its total uncertainty in the case of discrete distributions 
\begin{align}
    \underbrace{\mathbb{I}_{\omega \sim \hat{q}}\left[ \hat{a}, \omega \mid s \right]}_{\textrm{model uncertainty}} &= \
    \underbrace{\mathbb{H}\left[ \hat{a} \mid s \right]}_{\textrm{total uncertainty}} - 
    \underbrace{\mathbb{E}_{\omega \sim \hat{q}}\left[ \mathbb{H}\left[ \hat{a} \mid \omega, s \right] \right]}_{\textrm{(expected) behavioral uncertainty}} \\
    \textrm{where  }\underbrace{\mathbb{H}\left[ \hat{a} \mid \omega, s \right]}_{\textrm{(model-specific) behavioral uncertainty}} &=
    \underbrace{\mathbb{E}_{\hat{\pi} \sim \hat{p}_{\omega}}\left[\mathbb{H}\left[ \hat{a} \mid \hat{\pi}, s \right]\right]}_{\textrm{intrinsic behavioral uncertainty}} + 
    \underbrace{\mathbb{I}_{\hat{\pi} \sim \hat{p}_{\omega}}\left[ \hat{a}, \hat{\pi} \mid s  \right]}_{\textrm{extrinsic behavioral uncertainty}} 
    \label{eqn:model_uncertainty}
\end{align} or by directly computing $\textrm{Var}_{\omega \sim \hat{q}} \left[ \mathbb{E}\left[ \hat{a} \mid \omega, s \right] \right] \triangleq \textrm{Var}_{\omega \sim \hat{q}} \left[ \mathbb{E}_{\hat{\pi} \sim \hat{p}_{\omega}}\left[ \mathbb{E}\left[ \hat{a} \mid \hat{\pi}, s \right] \right] \right]$ in the case of continuous distributions.
As seen from these formulae, to estimate model uncertainty, we need to approximate two nested expectations: an outer expectation over $\hat{q}$ and an inner expectation over $\hat{p}_{\omega}$ (see Appendix B for more detail). 
This is highly computationally expensive to perform, especially on large training data and with a large number of teachers. 
Moreover, using a finite-sample Monte Carlo approximation of the expectation over $\hat{p}_{\omega}$ adds \textit{sampling uncertainty} to estimations of model uncertainty.
Assuming the sample size is fixed, this uncertainty correlates with the agent extrinsic uncertainty (i.e. the uncertainty of $\hat{p}_{\omega}$).
When learning from teachers that exhibit high extrinsic uncertainty, the agent extrinsic uncertainty increases over time, causing the estimations of model uncertainty to also grow.
This contradicts the nature of model uncertainty (it should decrease with more observed data).
Experimental results in \autoref{sec:exp_results} verify that if the approximation sample size is small, estimations of model uncertainty does \textit{increase} over time when learning with highly extrinsically uncertain teachers.

\section{Active Performance-Based Imitation Learning (\method)}
We propose directly using performance of the agent to drive query decisions.
Specifically, we base query decisions on how well the agent performs in comparison with the teachers. 
Let $d_T^{\star}$ be the expected final distance to the goal set $S_{\textrm{goal}}$ if the agent always queries the teachers (and follows their decisions) in every time step.
This quantity represents the average performance of the teachers on the main task. 
We define the \textit{performance gap}\footnote{We use performance gap instead of absolute performance to effectively deal with non-optimal teachers.
The agent's performance-matching goal remains unchanged regardless of the absolute performance of the teachers.} with respect to the teachers in a state $s$ as $g(s) \triangleq d(s) - d_T^{\star}$.
Also, let $s_{t}^{\mapsto t'}$ be the state the agent arrives in if it starts in state $s_t$ in time step $t$ and follows its own policies until time step $t'$.

At a high level, the goal of the agent is to close the performance gap at the end of a task episode (i.e. $\mathbb{E}\left[g\left(s_0^{\mapsto T}\right)\right] = 0$).
We train a query policy that predicts whether this goal can be achieved from a current state. 
If the goal cannot be reached, the agent queries the  teachers in the current state; otherwise, it relies on its own execution policy to proceed. 
In practice, closing the performance gap may be too demanding of the agent, especially at the early stage of training. 
We relax this goal by only requiring the agent to achieve \textit{substantial progress} in closing the gap.
The agent is said to achieve substantial progress in state $s_t$ if it meets one of the following conditions 
\begin{enumerate}
    \item the final performance gap is close to zero: $\mathbb{E} \left[g\left(s_t^{\mapsto T} \right)\right] \leq \epsilon$ for a small constant $\epsilon$;
    \item the performance gap will be reduced by at least a factor of $\sigma > 1$ between two future time steps: $\exists t \leq i < j:  \mathbb{E} \left[g\left(s_t^{\mapsto i} \right)\right] \geq \sigma \cdot \mathbb{E} \left[g\left(s_t^{\mapsto j} \right)\right]$, where $\sigma$ is a constant.
\end{enumerate} 

Let $\jointpol = \left(\exepol, \askpol \right)$ be the agent's current policies.
We define $\proset$ as the set of states from which the agent achieves substantial progress if it follows $\jointpol$ thereafter.
Given a fixed execution policy, we search for a query policy that optimizes the following objective
\begin{align}
    \min_{\askpol} \mathbb{E}_{s \sim P_{\jointpol}, \hat{a}^{\textrm{ask}}_s \sim \askpol(s)} \left[ \ell_{\textrm{apil}}\left(s, \jointpol, \hat{a}^{\textrm{ask}}_s \right) 
    \triangleq
    \mathds{1}\{ \hat{a}^{\textrm{ask}}_s = \texttts{query}\}^{\mathds{1}\left\{ s \in \proset \right\}}
    \right]
    \label{eqn:query_obj}
\end{align} This objective alternates between query minimization and performance maximization.
If the agent will make substantial progress from $s$ (i.e. $s \in \proset$),  $\ell_{\textrm{apil}} = \mathds{1}\{ \hat{a}^{\textrm{ask}} = \texttts{query}\}$. The objective minimizes the expected query count in $s$. 
Otherwise,  $\ell_{\textrm{apil}} = 1$. The agent receives a constant loss in $s$ regardless of its query decision.
To minimize this loss, the agent has to improve the performance of $\jointpol$ so that the probability of visiting $s$ diminishes or $s$ becomes a progressable state ($s \in \proset$).
Concretely, the objective can be rewritten as a sum of two sub-objectives
\begin{align}
    \min_{\askpol} \ \underbrace{\sum_{s^{\odot} \in \proset} P_{\jointpol}(s^{\odot}) \cdot \askpol(\texttts{query} \mid s^{\odot})}_{\textrm{minimized by reducing query rate}} \ \ \ + \underbrace{\sum_{s^{\otimes} \notin \proset} P_{\jointpol}(s^{\otimes})}_{\textrm{minimized by improving performance}}
\end{align} We teach the query policy imitate two \textit{opposing} query teachers, each of which minimizes only  one sub-objective. Let $\delta_a$ is a deterministic policy that always selects action $a$. 
It is easy to see that first sub-objective (alone) is minimized when $\askpol = \delta_{\texttts{continue}}$, while the second sub-objective
(alone) is minimized when $\askpol = \delta_{\texttts{query}}$.
We combine these two teachers into a single teacher policy  $\askexp(\texttts{query} \mid s) = \mathds{1}\{ s \notin \proset \}$ and teach $\askpol$ to imitate this policy.
The decision-making capabilities of the agent controls the balance between the two query teachers.
At the beginning of training, the agent cannot make substantial progress in most states, thus $\delta_{\textrm{query}}$ dominates, helping the agent to improve quickly. 
As the agent becomes more capable, the influence of $\delta_{\textrm{continue}}$ amplifies, gradually reducing the query rate.  
If the performance of the agent drops due to reducing queries, the performance-maximization sub-objective enlarges and the agent will be taught to increase the query rate to regain performance.
The competition between the two teachers eventually settles at an equilibrium where the agent reaches maximal performance or learning capacity, while issuing a minimal number of queries.


\noindent \textbf{Implementation}. \autoref{alg:apil} details our algorithm \method. Deciding whether a state $s$ is in $\proset$ is intractable because of the expectation over all possible trajectories starting from $s$.
In practice, for every training episode, we only generate a single trajectory $\traj = \left\{ \left( s_t, \exea, \aska \right)\right\}_{t = 1}^T$ using $\jointpol$, and only consider progress with respect to this trajectory.
Let $\appoxproset$ be a set containing states $s_t \in \traj$ that satisfy (a) $t = T$ and $g(s_T) \leq \epsilon$ or (b) there exists $i, j$ such that $t \leq i < j$, $\hat{a}_{i}^{\textrm{ask}} = \hat{a}_{j}^{\textrm{ask}} = \texttts{query}$, and $ g(s_i) \geq \sigma \cdot g(s_j)$. We substitute $\proset$ with $\appoxproset$ when computing the query reference actions $\askaref$. 

\section{Experimental Setup}
\begin{figure}[t!]
\begin{minipage}{0.45\textwidth}
\centering
\begin{subfigure}[t]{.4\textwidth}
        \centering
        \includegraphics[width=0.8\textwidth]{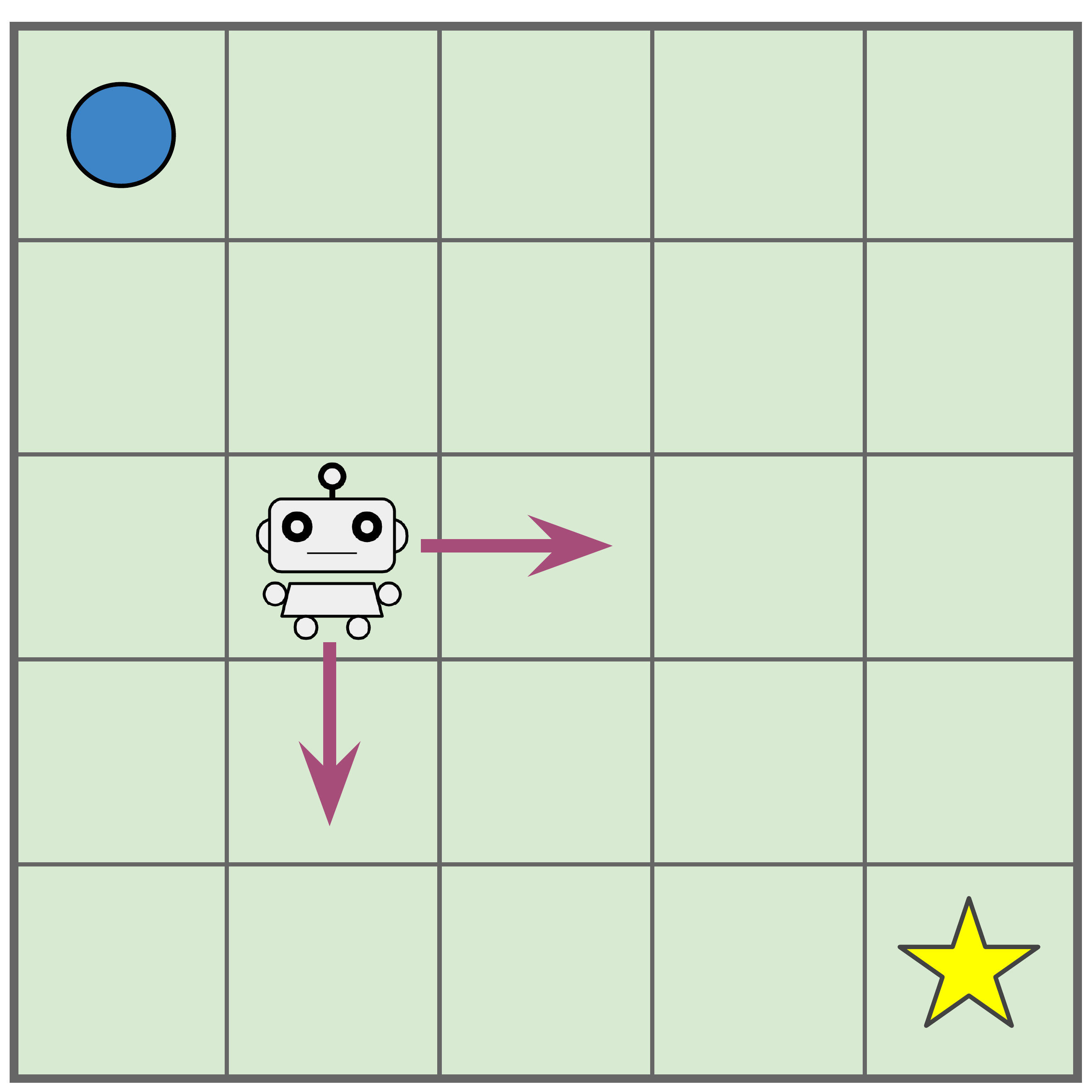}
        \captionof{figure}{GRID}
        \label{fig:grid}
    \end{subfigure}%
\begin{subfigure}[t]{0.6\textwidth}
        \centering
        \includegraphics[width=.9\textwidth]{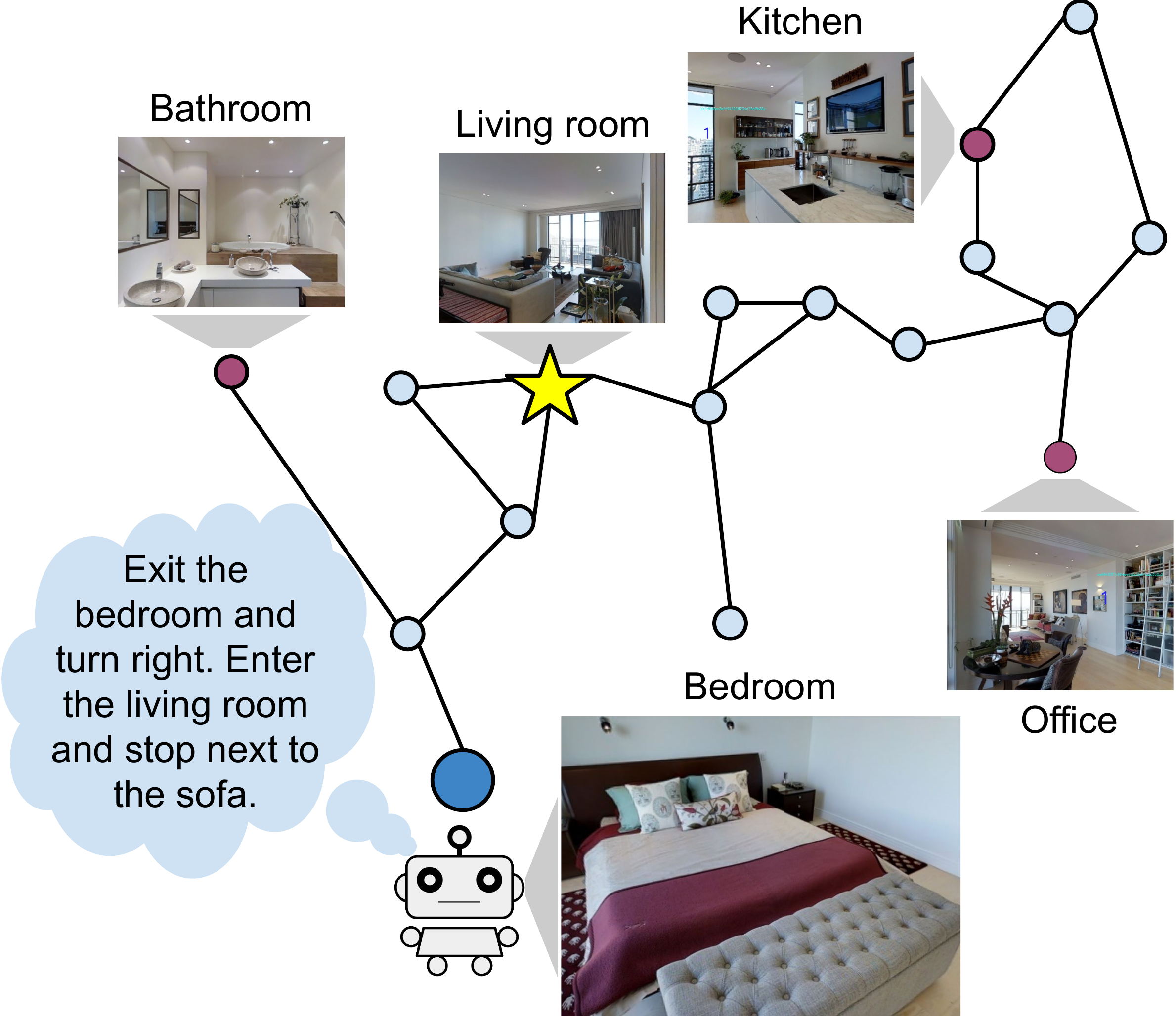}
        \captionof{figure}{R2R}
        \label{fig:r2r}
    \end{subfigure}%
\captionof{figure}[Tasks and environments: (a) grid-world navigation where the agent always succeeds no matter what it does and (b) photorealistic navigation by following language instructions.]{Tasks and environments (\circleB : start location, \starEnd : goal location): (a) grid-world navigation where the agent always succeeds no matter what it does, (b) photorealistic navigation by following language instructions.}
\end{minipage}
\hfill
\begin{minipage}{0.5\textwidth}
\centering
\includegraphics[width=\textwidth]{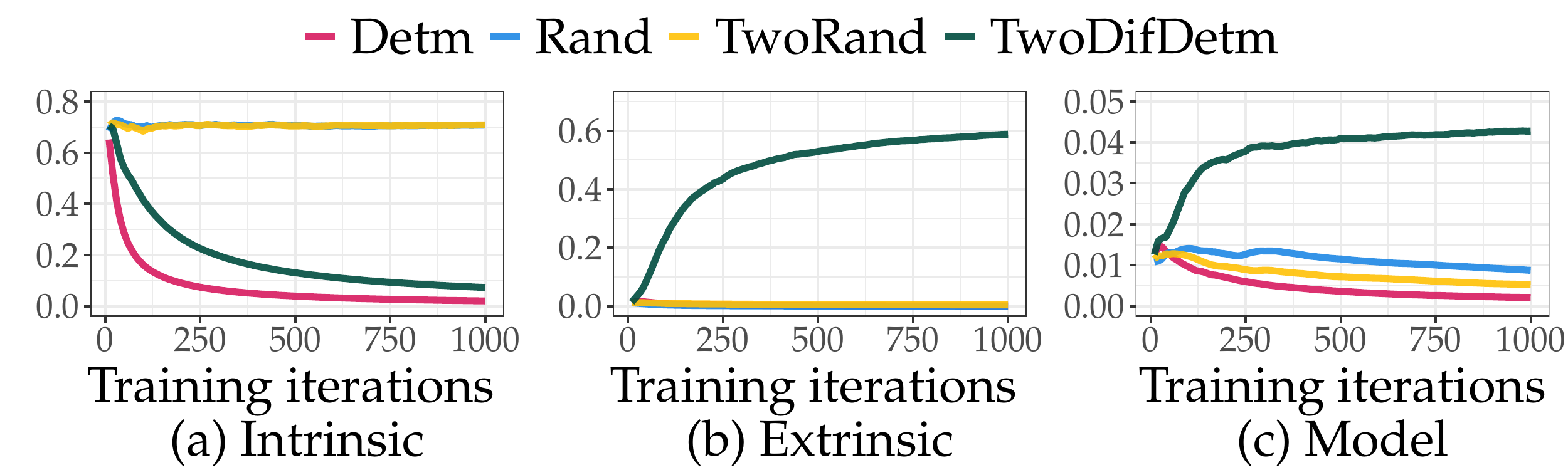}
\vspace{-0.8cm}
\captionof{figure}{Progress (during training) of the agent (approximate) uncertainty when learning the GRID task from different teacher models.}
\label{fig:grid_uncertainty}
\vskip\baselineskip
\vspace{-0.3cm}
\centering
\includegraphics[width=\textwidth]{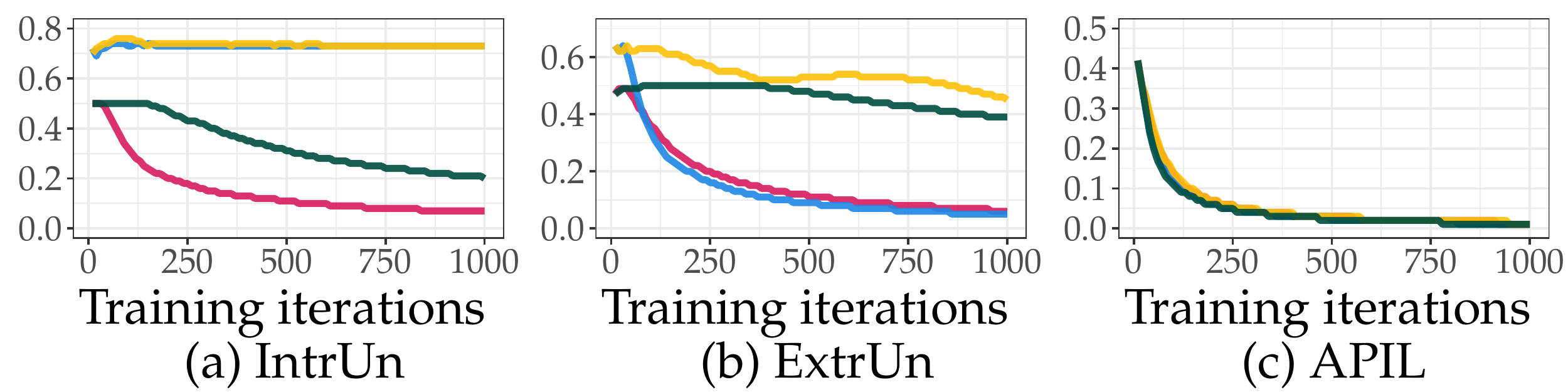}
\vspace{-0.8cm}
\captionof{figure}{Query rate (during training) of various query policies when learning the GRID task from different teacher models. }
\label{fig:grid_method}
\end{minipage}
\vspace{-0.5cm}
\end{figure}

\noindent\textbf{Tasks and environments}. We conduct experiments on two problems: grid-world navigation (\toynav) and instruction-following navigation in photo-realistic environments (\rr). 
In \toynav (\autoref{fig:grid}), an agent wants to go to the bottom-right cell from the top-left cell in a 5$\times$5 grid world.
In each step, the agent can either go right or down one cell, but it cannot escape the grid.
There are no obstacles in the environment. 
Hence, no matter how the agent behaves, it will always successfully complete the task after eight steps. 
In \rr (\autoref{fig:r2r}), an agent follows a natural language instruction to reach a pre-determined location. 
The environments are implemented by the Matterport3D simulator \cite{anderson2018vision,chang2017matterport3d}, which photo-realistically emulates the view of a person walking inside residential buildings. 
We use the \rr instruction dataset collected by \cite{anderson2018evaluation}.
Walking in an environment is modeled as traversing on a undirected graph, with edges connecting nearby locations. 
The agent is given the panoramic view of its current location.
In every time step, it either stops or walks to a location that is adjacent to its current location on an environment graph.
The agent successfully follows an instruction if in the end it stands within three meters of the goal location.
R2R is a highly complex sequential decision-making problem, requiring the agent to learn visually grounded representation of the language instructions to identify landmarks and execute actions referred to in those instructions in the environments. 

\noindent\textbf{Teacher models}. We implement four teacher models. Let $A^{\textrm{ref}}_s$ be an ordered set of reference actions in state $s$. A \detm teacher always references the first action in this set, whereas a \rand teacher uniformly randomly recommends an action in the set. 
Based on these teachers, we construct two types of teacher committee. A \tworand teacher consists of two \textit{identical} \rand teachers. 
A \twodifdetm teacher mixes two \textit{different} \detm teachers, where one teacher always selects the first action in $A^{\textrm{ref}}_s$ and the other always selects the last action.
With these committees, a member teacher is selected with a probability of $0.5$ at the \textit{beginning} of each training episode and will interact with the agent throughout the entire episode.
The "Expectation" column of \autoref{tab:uncertainty} highlights the differences in the intrinsic and extrinsic uncertainty of the teacher models. 
Note that the \tworand teacher has zero extrinsic uncertainty because its member teachers are identical. 
See Appendix A on how we implement the teachers in each problem.

\noindent\textbf{Query policy baselines}. We compare \method with behavior cloning (\bc) and \dgger \cite{ross2010efficient}, two standard approaches to \iil which query the teachers in every time step.
We implement common \ail methods: (a) \intrinsic (intrinsic behavioral uncertainty-based), (a) \extrinsic (extrinsic behavioral uncertainty-based), (b) \behave (total behavioral uncertainty-based), and (c) \errpred (similar to \cite{zhang2017query}, queries when predicting that the margin $1 - \exepol\left(a_{s}^{\textrm{exe}\star} \mid s\right)$ exceeds a threshold).
For \method, we set $\sigma = 2$, $\epsilon = 0$ in \toynav, and $\epsilon = 3$ in \rr.
The hyperperameters of other methods are tuned so that their query rates are on par with that of \method when learning with the \detm teacher. 
Then their hyperparameters are then kept the same in all experiments on a problem. 
See Appendix B and C for the detailed implementation of the query policies and the agent models. 

\section{Results}
\label{sec:exp_results}
\noindent\textbf{Does the teacher persona-aware model learn the teacher models?} We verify if the teacher persona-aware model (\autoref{sec:mind}) captures the characteristics of the teacher models. 
\autoref{tab:uncertainty} shows that the intrinsic and extrinsic uncertainty learned by our agents closely match our expectations. 
The intrinsic uncertainty learned from teachers with deterministic members (\detm, \twodifdetm) is lower than learned from those with non-deterministic members (\rand, \tworand).
The model successfully recognizes that the two members of \tworand are similar, exhibiting near-zero extrinsic uncertainty for this teacher. 
On other hand, the non-zero extrinsic uncertainty learned from the \twodifdetm teacher demonstrates that the model is able to identify the differences between the two member teachers.  
\autoref{fig:grid_uncertainty} offers more insights into how each type of uncertainty progresses during training.

\noindent\textbf{Behavior of model uncertainty approximation}. 
We observe that the temporal patterns of the approximate model uncertainty (\autoref{fig:grid_uncertainty}a) are similar to those of the approximate extrinsic uncertainty (\autoref{fig:grid_uncertainty}b).
This shows that the approximate model uncertainty reflects the characteristics of extrinsic uncertainty rather than those of model uncertainty.
This phenomenon is anticipated by our theory (\autoref{sec:uncertainty}): the approximate model uncertainty is inflated by sampling uncertainty that strongly correlates with extrinsic uncertainty. 
In Fig \ref{fig:model_inflat}, we plot the approximate model uncertainty when learning the GRID task from the \twodifdetm teacher.
We vary the number of samples used to approximate the expectation over $\hat{p}_{\omega}$ in the formula of model uncertainty (\autoref{eqn:model_uncertainty}). 
Results show that, as the number of samples increases, the sampling uncertainty decreases and the influence of extrinsic uncertainty diminishes. 
With 50 samples, the approximate model uncertainty starts behaving like the true model uncertainty (i.e. it decreases as training progresses). 
This shows that being able to estimate model uncertainty reliably comes with tremendous computational cost. 

\noindent\textbf{Efficiency and robustness of \method}. 
In both tasks, \method consistently significantly reduces the number of queries to the teachers. 
In the \toynav task, the optimal query behavior is to not query at all, regardless of the teacher. 
As seen from \autoref{fig:grid_method}, \method is the only method that correctly learns this behavior for all teacher models. 
In contrast, \intrinsic and \extrinsic are misled by the high intrinsic and extrinsic  uncertainty (respectively) acquired by copying the teacher behaviors. 
\method also outperforms baseline methods on the more challenging \rr task, in terms of both success rate and query rate (\autoref{tab:r2r_results}).
It achieves even higher success rate than \bc and \dgger in the cases of \tworand and \twodifdetm.
The method reduces the query rate by three times while only slightly raising the query rate by $\sim$2\% in the face of uncertain teachers. 
In contrast, \intrinsic and \behave increases their query rates by $\sim$8\% when learning from \rand and \tworand, which have high intrinsic uncertainty.

\begin{table}[t!]
\begin{minipage}{0.42\textwidth}
\centering
\includegraphics[width=0.7\textwidth]{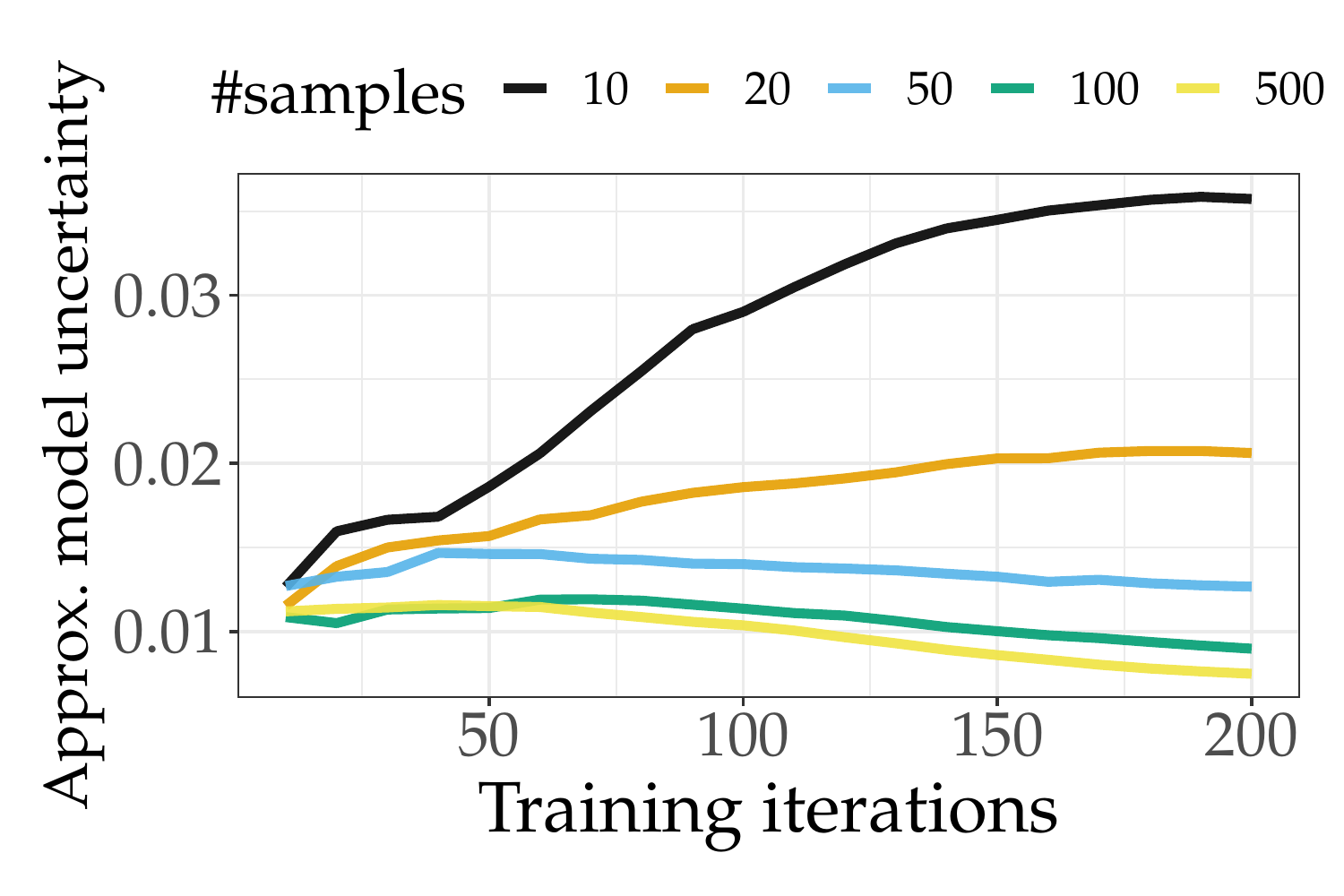}
\captionof{figure}{Approximate model uncertainty when learning the GRID task from the \textsc{TwoDifDetm} teacher. Increasing the number of samples used for approximating the expectation over $\hat{p}_{\omega}$ reduces the influence of extrinsic uncertainty.}
\label{fig:model_inflat}
\end{minipage}
\hfill
\begin{minipage}{0.55\textwidth}
\centering
\scriptsize
\captionof{table}{Agent behavioral uncertainty learned from different teacher models. The "Expectation" column shows the intrinsic and extrinsic uncertainty if the agent perfectly learned each teacher model.}
\begin{tabular}{lcccccc}
	\toprule
	& \multicolumn{2}{c}{Expectation} 
	& \multicolumn{2}{c}{GRID} 
	& \multicolumn{2}{c}{R2R} \\ \cmidrule(r){2-3}  \cmidrule(r){4-5} \cmidrule(r){6-7}
	Teacher & Intrinsic & Extrinsic & Intrinsic & Extrinsic & Intrinsic & Extrinsic \\ \cmidrule(r){1-1} \cmidrule(r){2-2} \cmidrule(r){3-3} \cmidrule(r){4-4} \cmidrule(r){5-5}
	\cmidrule(r){6-6} \cmidrule(r){7-7}
	Detm & 0 & 0 & 0.04 & 0.00 & 0.21 & 0.00 \\
	Rand & > 0 & 0 & 0.72 & 0.00 & 0.27 & 0.00 \\
	TwoRand & > 0 & 0 & 0.72 & 0.00 & 0.27 & 0.00 \\
	TwoDifDetm & 0 & > 0 & 0.05 & 0.56 & 0.26 & 0.05 \\
    \bottomrule 
\end{tabular}
\label{tab:uncertainty}
\end{minipage}
\vspace{-0.6cm}
\end{table}

\begin{table}[t!]
\centering
\scriptsize
\caption{Success rate and query rate (on the R2R test set) of query policies when learning from different teachers. Each method uses the same hyperparameters for all teachers. We do not report results of \extrinsic because we could not find a working threshold for this method.}
\label{tab:r2r_results}
\centering
\begin{tabular}{lcccccccc}
	\toprule
	& \multicolumn{4}{c}{Success rate (\%) $\uparrow$}
	& \multicolumn{4}{c}{Query rate (\%) $\downarrow$}  \\
	\cmidrule(r){2-5} \cmidrule(r){6-9}
	Method& Detm & Rand
	& TwoRand & TwoDifDetm
	& Detm & Rand
	& TwoRand & TwoDifDetm  \\ 
	\cmidrule(r){1-1} 
	\cmidrule(r){2-2}
	\cmidrule(r){3-3}
	\cmidrule(r){4-4}
	\cmidrule(r){5-5}
	\cmidrule(r){6-6}
	\cmidrule(r){7-7}
	\cmidrule(r){8-8}
	\cmidrule(r){9-9}
	BC & 30.2 & 32.2 & 30.1 & 30.1 & 100.0 & 100.0 & 100.0 & 100.0 \\
	DAgger & \textbf{35.2} & \textbf{35.1} & 32.8 & 34.5 & 100.0 & 100.0 & 100.0 & 100.0  \\
	IntrUn & 29.2 & 30.4 & 33.9 & 33.0 & ~~30.5 & ~~38.9 & ~~38.3 & ~~32.4 \\
	BehvUn & 29.6 & 29.8 & 32.2 & 33.8 & ~~27.1 & ~~39.0 & ~~38.4 & ~~34.5 \\
	ErrPred & 34.2	& 34.9 & 33.1 &	31.0 & ~~29.0 & ~~31.1 & ~~30.9 & ~~32.2 \\
	APIL (ours) & \textbf{35.2} & 34.6 & \textbf{36.9} & \textbf{35.7} & ~~\textbf{27.8} & ~~\textbf{29.6} & ~~\textbf{29.7} & ~~\textbf{29.6} \\
    \bottomrule
\end{tabular}
\vspace{-0.3cm}
\end{table}

\section{Related work and Conclusion}
Imitation learning with multiple, non-deterministic teachers has previously been studied but the agent is modeled as a single policy \cite{li2018oil}. 
Learning all modes of the teacher policies is a challenging problem due to the characteristics of current learning paradigms \cite{ke2019imitation}.
\cite{li2017infogail} present a solution by introducing a latent variable, whose distribution is learned in an unsupervised manner with an mutual-information maximization objective.
Assuming access to the teacher true identities, we develop a general framework for learning \textit{all} teacher policies by modeling a distribution over an entire policy space. 
To our knowledge, we are the first to study teacher behavioral uncertainty in active imitation learning. 
We highlight challenges in applying traditional uncertainty-based approaches \cite{gal2017deep,houlsby2011bayesian} to this problem, showing that different types of the agent uncertainty become misleading signals for making query decisions.  
To overcome those challenges, it is necessary that the teachers provide additional signals to help the agent monitor and predict its progress on a task. 
In this work, we employ a simple, sparse signal: the distance to the goal that is only given upon request. 
However, this signal can even be difficult to estimate accurately in real-world scenarios.
In future work, we will evaluate the robustness of our active learning algorithm under noisy, erroneous estimations of the distance metric. 
Another limitation of the algorithm is that currently the agent is not able to determine whether it has reached maximal learning capacity.
We will investigate directions to enabling the agent to track its progress over an entire course of learning rather than within only a training episode.

\bibliography{journal_full, neurips_2020}
\bibliographystyle{named}

\clearpage
\section*{Appendices}
\setcounter{section}{0}
\renewcommand\thesection{\Alph{section}}



\section{Simulating Teacher Uncertainty}

We describe how we construct the (ordered) reference action set $A^{\textrm{ref}}_s$ in each problem.
In \toynav, $A^{\textrm{ref}}_s$ contains one valid action ("up" or "down") in the nine edge cells , and two valid actions ("up" and "down") in the 16 inner cells .
In \rr, each action corresponds to a location that is adjacent to the agent's current location on the environment graph. 
Let $s$ and $s'$ be two adjacent locations on the ground-truth path corresponding to an instruction.
We construct a set of locations $A^{\textrm{augmt}}$ such that, for every $s_{\textrm{augmt}} \in A^{\textrm{augmt}}$, $s_{\textrm{augmt}} \neq s'$, $s_{\textrm{augmt}}$ is adjacent to $s$ and the angle between two segments $(s, s')$ and $(s, s_{\textrm{augmt}})$ is less than or equal to 60 degrees.
We then define $A^{\textrm{ref}}_s \triangleq \{s'\} \cup A^{\textrm{augmt}} $.
On average, $|A^{\textrm{ref}}_s| = 1.6$ in this problem.

\section{Approximating Agent Uncertainty}

Because the action spaces of our problems are discrete, we use Shannon entropy to measure uncertainty. 
All policies refer in this section are execution policies. 
For brevity, we will drop the subscript "exe" in the policy notations. 
Each type of agent uncertainty is defined based on the following decomposition provided in the paper: 
\begin{align}
    \underbrace{\mathbb{I}_{\omega \sim \hat{q}}\left[ \hat{a}, \omega \mid s \right]}_{\textrm{model uncertainty}} &= \
    \underbrace{\mathbb{H}\left[ \hat{a} \mid s \right]}_{\textrm{total uncertainty}} - 
    \underbrace{\mathbb{E}_{\omega \sim \hat{q}}\left[ \mathbb{H}\left[ \hat{a} \mid \omega, s \right] \right]}_{\textrm{(expected) behavioral uncertainty}} \\
    \textrm{where  }\underbrace{\mathbb{H}\left[ \hat{a} \mid \omega, s \right]}_{\textrm{(model-specific) behavioral uncertainty}} &=
    \underbrace{\mathbb{E}_{\hat{\pi} \sim \hat{p}_{\omega}}\left[\mathbb{H}\left[ \hat{a} \mid \hat{\pi}, s \right]\right]}_{\textrm{intrinsic behavioral uncertainty}} + 
    \underbrace{\mathbb{I}_{\hat{\pi} \sim \hat{p}_{\omega}}\left[ \hat{a}, \hat{\pi} \mid s  \right]}_{\textrm{extrinsic behavioral uncertainty}} 
\end{align}

Let $\hat{\pi}_{\omega}(\cdot \mid s) \triangleq \mathbb{E}_{\hat{\pi} \sim \hat{p}_{\omega}}\left[ \hat{\pi}(\cdot \mid s) \right]$, and $\hat{\pi}(\cdot \mid s) \triangleq \mathbb{E}_{\omega \sim \hat{q}}\left[ \hat{\pi}_{\omega}(\cdot \mid s) \right]$. 
The approximations of these policies are computed via Monte Carlo sampling:
\begin{align}
    \hat{\pi}_{\omega}(\cdot \mid s) &\approx \tilde{\pi}_{\omega}(\cdot \mid s) = \frac{1}{N_1} \hat{\pi}^{(i)}_{\omega}(\cdot \mid s) \textrm{ where } \hat{\pi}^{(i)}_{\omega} \sim \hat{p}_{\omega}(\cdot \mid s) \\
    \hat{\pi}(\cdot \mid s) &\approx \tilde{\pi}(\cdot \mid s) = \frac{1}{N_2} \tilde{\pi}_{\omega^{(i)}}(\cdot \mid s) \textrm{ where } \omega^{(i)} \sim \hat{q}
\end{align}

Then, each type of uncertainty is calculated as follows:
\begin{align}
    \textrm{Total uncertainty} &\approx \mathbb{\tilde{H}}\left[ \hat{a} \mid s \right] = -\sum_{a} \tilde{\pi}\left(a \mid s \right)\log \tilde{\pi}\left(a \mid s \right) \\
    \textrm{Model-specific behavioral uncertainty} &\approx \mathbb{\tilde{H}}\left[ \hat{a} \mid \omega, s \right] = -\sum_{a} \tilde{\pi}_{\omega} \left(a \mid s \right)\log \tilde{\pi}_{\omega} \left(a \mid s \right) \\
    \textrm{Intrinsic behavioral uncertainty} &\approx \mathbb{\tilde{E}}_{\hat{\pi} \sim \hat{p}_{\omega}}\left[ \mathbb{H}\left[ \hat{a} \mid \hat{\pi}, s\right] \right] \\
    &= \frac{1}{N_1}\left[ -\sum_{a} \hat{\pi}^{(i)}_{\omega}\left(a \mid s \right)\log \hat{\pi}^{(i)}_{\omega}\left(a \mid s \right) \right] \\
    & \textrm{where } \hat{\pi}^{(i)}_{\omega} \sim \hat{p}_{\omega}(\cdot \mid s) \\
    \textrm{Extrinsic behavioral uncertainty} &\approx \mathbb{\tilde{H}}\left[ \hat{a} \mid \omega, s \right] - \mathbb{\tilde{E}}_{\hat{\pi} \sim \hat{p}_{\omega}}\left[ \mathbb{H}\left[ \hat{a} \mid \hat{\pi}, s\right] \right] \\
    \textrm{Model uncertainty} &\approx  \mathbb{\tilde{H}}\left[ \hat{a} \mid s \right] - \frac{1}{N_2}\left[ \mathbb{\tilde{H}}\left[ \hat{a} \mid \omega^{(i)}, s \right] \right] \\
    & \textrm{where } \omega^{(i)} \sim \hat{q}
\end{align}

Unless further specified, we set $N_1 = 5$ and $N_2 = 10$ in all experiments.

\section{Model architecture and training configuration}

\noindent\textbf{Model architecture}. 
In \toynav, the identity model $\rho_{\psi}$, the persona-conditioned execution policy $\hat{\pi}_{\theta, h}$, and the query policy $\askpol$ are feed-forward neural networks with one hidden layer of size 100. The size of the persona embeddings $\phi$ is 50. 
In \rr, the identity model $\rho_{\psi}$ is a feed-forward neural network with one hidden layer of size 512. 
The size of the persona embeddings is 256. 
The persona-conditioned execution policy is a linear model.
The query policy is an LSTM-based neural network with a hidden size of 512. 

We model the parameter posterior distribution $\hat{q}$ by applying MC dropout \cite{gal2016dropout} with a dropout rate of 0.2 to the input of the persona-conditioned policy.

\noindent\textbf{Input features}. The persona-conditioned execution policy takes the concatenation of a state representation $s$ and a persona embedding $h$ as input. 
In \toynav, $s$ is the raw representation of the grid with empty cells, the agent's current location, and the goal location encoded by different float numbers in $[0, 1]$. 
In \rr, we employ the transformer-based model described in \cite{nguyen2019hanna} to compute the state representations $s$. 
The input of the query policies contains three features: (a) a current state representation $s$, (b) an action distribution of the approximate mean execution policy $\tilde{\pi}_{\textrm{exe}}(\cdot \mid s)$ (whose computation is described in the previous section) (c) an embedding of the number of remaining time steps $T - t$, where $T$ is the time constraint and $t$ is the current time step.

\noindent\textbf{Training}. We train all models with the Adam optimizer \cite{kingma2014adam}. 
In \toynav, we train the agent for 1000 iterations with a batch size of 1 and a learning of $10^{-3}$.
In \rr, we train the agent for $10^5$ iterations with a batch size of 100 and a learning of $10^{-4}$.

\end{document}